\documentclass[10pt,twocolumn,letterpaper]{article}

\usepackage{cvpr}              %

\usepackage{graphicx}
\usepackage{amsmath}
\usepackage{amssymb}
\usepackage{booktabs}
\usepackage[normalem]{ulem}
\usepackage{paralist}
\usepackage{microtype}
\usepackage{soul}
\usepackage{xcolor}
\usepackage{hhline}
\usepackage{adjustbox}
\usepackage{multirow}
\usepackage{upgreek}
\usepackage{pifont}
\usepackage{amsfonts}
\usepackage{bbold}
\usepackage{enumitem}
\newcommand{\cmark}{\text{\ding{51}}}
\newcommand{\xmark}{\text{\ding{55}}}
\newcommand{\walon}[1]{\textcolor{black}{#1}}

\newcommand{\students}[1]{\textcolor{black}{#1}}

\usepackage[pagebackref,breaklinks,colorlinks,citecolor=blue,linkcolor=blue]{hyperref}

\usepackage[capitalize]{cleveref}
\crefname{section}{Sec.}{Secs.}
\Crefname{section}{Section}{Sections}
\Crefname{table}{Table}{Tables}
\crefname{table}{Tab.}{Tabs.}

\def\cvprPaperID{1909} %
\def\confName{CVPR}
\def\confYear{2022}

\begin{document}

\title{Self-Supervised Feature Learning from Partial Point Clouds \\ via Pose Disentanglement}

\author{Meng-Shiun Tsai$^{\ast1}$\quad Pei-Ze Chiang$^{\ast1}$\quad Yi-Hsuan Tsai$^2$\quad Wei-Chen Chiu$^{1}$ \\
$^1$National Yang Ming Chiao Tung University, Taiwan \quad$^2$NEC Labs America\\
}
\maketitle

\begin{abstract}
   Self-supervised learning on point clouds has gained a lot of attention recently, since it addresses the label-efficiency and domain-gap problems on point cloud tasks. In this paper, we propose a novel self-supervised framework to learn informative representations from partial point clouds. We leverage partial point clouds scanned by LiDAR that contain both content and pose attributes, and we show that disentangling such two factors from partial point clouds enhances feature representation learning. To this end, our framework consists of three main parts: 1) a completion network to capture holistic semantics of point clouds; 2) a pose regression network to understand the viewing angle where partial data is scanned from; 3) a partial reconstruction network to encourage the model to learn content and pose features. To demonstrate the robustness of the learnt feature representations, we conduct several downstream tasks including classification, part segmentation, and registration, with comparisons against state-of-the-art methods. Our method not only outperforms existing self-supervised methods, but also shows a better generalizability across synthetic and real-world datasets.
\end{abstract}
\let\thefootnote\relax\footnote{$\ast$ Both authors contributed equally to the paper}

\vspace{-2mm}
\section{Introduction}
\label{sec:introduction}

\begin{figure}[ht]
    \begin{center}
        \includegraphics[width=1\columnwidth]{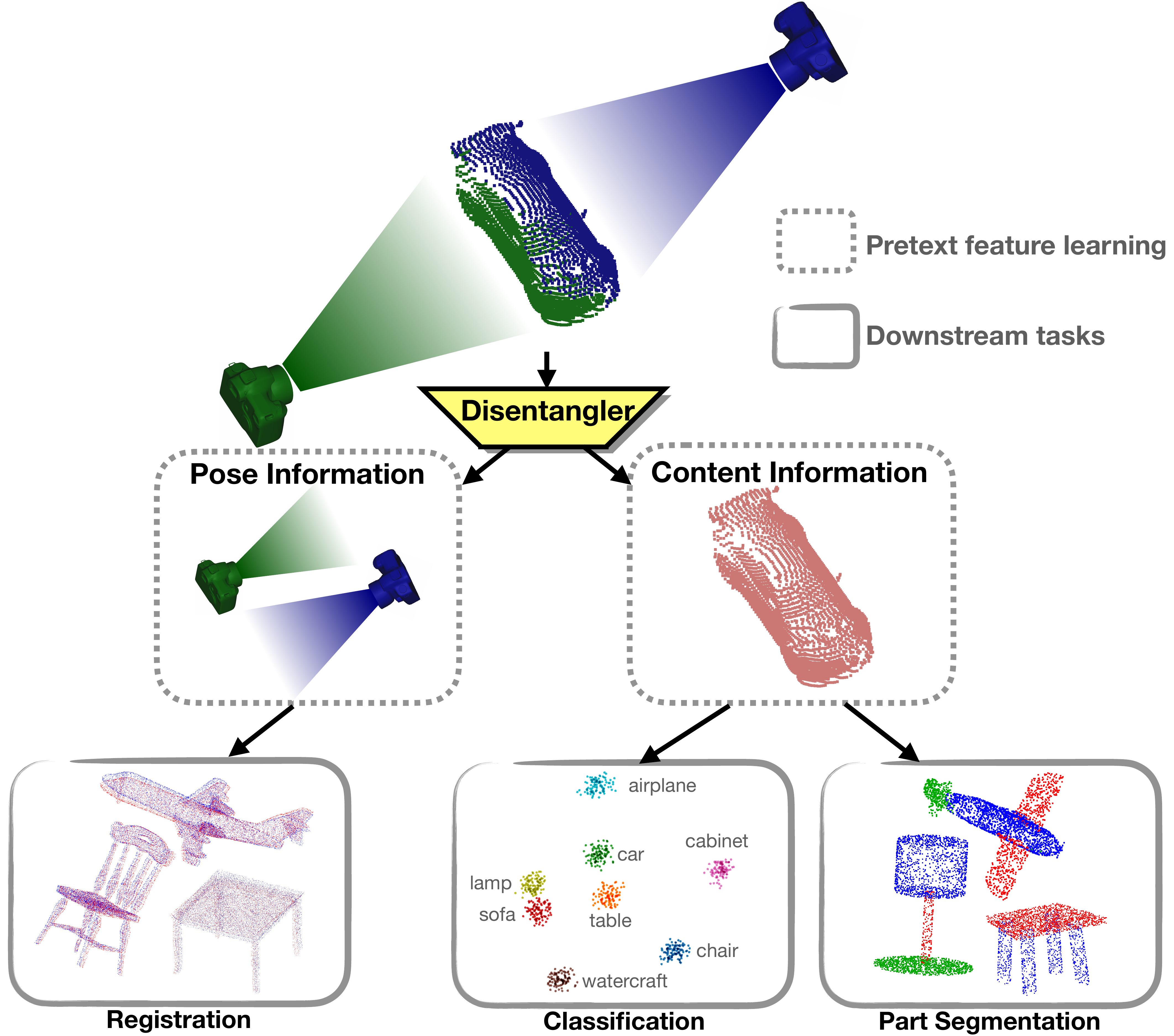}
    \end{center}
    \vspace{-2mm}
    \caption{
    We propose to learn the feature representation of 3D point cloud data in a self-supervised manner via point cloud completion. Our model learns to disentangle the feature into the content and pose parts, where the former is beneficial for the downstream tasks of classification and part segmentation, while the latter helps to improve the task of point cloud registration.}
    \label{fig:teaser}
    \vspace{-1em}
\end{figure}

Point clouds provide one of the most intuitive representations for 3D object models, and they are extensively adopted as the data format for the recognition tasks in different application scenarios, such as autonomous vehicles, robotics, and architectonics. Recently, with the rapid development of deep learning techniques, various network architectures are proposed for learning effective features of the point cloud data, e.g., PointNet~\cite{qi2017pointnet}, PointNet++~\cite{qi2017pointnet++}, and DGCNN~\cite{dgcnn}.
However, the success of learning the recognition models based on these feature backbones typically relies on the large-scale supervised dataset, in which it could be quite expensive to manually collect the ground truth labels and the learnt feature representations of point cloud data are less generalizable across different tasks.

These potential issues motivate various research development in learning effective feature representations from point clouds via unsupervised or self-supervised manner, such as solving jigsaw puzzles~\cite{alliegro2020joint, sauder2019self}, contrastive learning~\cite{rao2020global, xie2020pointcontrast}, and local structure modeling~\cite{han2019multi, hassani2019unsupervised}.
In this paper, we aim to seek for a different solution that is also intuitive and suitable for the point cloud data.
Inspired by the success of feature learning in the image~\cite{pathak2016context} and the natural language~\cite{vaswani2017attention} domains, which have been developed for years, we adopt an analogous strategy of image inpainting~\cite{pathak2016context}, but in a 3D point cloud version via \textit{learning to complete} the partial point cloud of a 3D object.
As such, similar benefits for feature learning as shown in image inpainting~\cite{pathak2016context} can be achieved by point cloud completion, which not only needs the semantic understanding of the point cloud data, but also learns better holistic feature representations via reconstructing the plausible missing parts.

However, simply completing the partial point cloud, as in the case of image domain, may not suffice the need for learning effective feature representations, as the point cloud data has different characteristics from images.
That is, given a 3D object point cloud, there could be multiple angles to view this data, such that each angle produces a different partial point cloud (in the top of Figure \ref{fig:teaser}).
Therefore, although the final completed point cloud is the same for all the partial point clouds obtained from the same object, the completion network may require to learn a different feature representation for each view-angle during the completion process. To take view-angle into consideration, instead of using existing point cloud completion frameworks~\cite{liu2020morphing, yuan2018pcn}, which only produce one feature representation that \textit{entangles} both the view-angle information and the content cue for the point cloud data, we propose to \textit{disentangle} these two factors and learn more effective feature representations in a self-supervised manner (see Figure \ref{fig:teaser}).

Specifically, we utilize two encoders to extract the content and pose (i.e., view-angle) features individually from each partial point cloud input. Then, in addition to performing completion using the content feature, the pose feature should be able to predict the pose of the input data and guide the reconstruction in specific view-angle.
Thus, we introduce another two modules: 1) a pose regression network to predict the view-angle of the partial point cloud using the pose feature; and 2) a partial reconstruction network to recover a specific partial point cloud, through the combination of a content feature from a view-angle $i$ and a pose feature from another view-angle $j$. As a result, no matter which view-angle the content feature is extracted from, the partial reconstruction should be mainly guided by the pose feature.
Therefore, our framework encourages the content and pose features to be more compact on themselves while being more disentangled to each other, which better facilitates the feature representation learning process.

We conduct extensive experiments to show the effectiveness of our self-supervised framework: 1) our learned content feature is able to provide favorable performance against state-of-the-art methods on the downstream tasks of classification and part segmentation, and 2) our learned pose features contribute to the pose-relevant downstream task of point cloud registration. The main contributions of our work are summarized as follows:

\begin{itemize}[noitemsep, topsep=0pt, partopsep=0pt]
    \item We propose a self-supervised framework based on point cloud completion for learning the feature representations from the partial point cloud data. 
    \item We develop a pipeline to disentangle point cloud feature representations into the content and pose factors, which enables the model to learn effective features.
    \item We show that the learnt content and pose features improve several downstream tasks (i.e., classification, part segmentation, and registration), while showing the generalizability across synthetic and real-world datasets.
    
\end{itemize}

\section{Related Work}
\label{sec:related}

\paragraph{Supervised Learning for Point Cloud Data.}
Due to the popularity of 3D scanning technologies and the advancement of deep learning techniques, learning to extract features from the point cloud data for recognition tasks is one of the active research topics. One early attempt via deep networks is PointNet~\cite{qi2017pointnet} which uses the max-pooling operation to aggregate the point-wise features into the global one, thus achieving the permutation invariant property. While PointNet neglects the local structure between neighboring points, its successor, PointNet++~\cite{qi2017pointnet++}, adopts the hierarchical neural network to progressively extract the features of a point cloud from multiple resolutions. 
Furthermore, DGCNN~\cite{dgcnn} leverages the graph neural network~\cite{kipf2016semi} to process the nearby points and their edges via the k-nearest neighbor algorithm in each of the intermediate feature spaces.
Other methods that also try to capture the relationships among the local regions in point clouds are developed in~\cite{li2018pointcnn, liu2019relation, xu2018spidercnn}. 

However, all the aforementioned works are based on the complete point cloud with supervised information, which is costly to collect in terms of time and expense. 
Even when we are able to use the synthetic dataset where the supervision is easier to obtain, the feature learnt from the synthetic data could be less generalizable to the real-world data due to the domain gap~\cite{angelina2019revisiting}.
In this work, we thereby focus on the self-supervised learning scheme to learn effective representations that can generalize better across datasets.

\begin{figure*}[!t]
    \begin{center}
        \includegraphics[width=0.9\textwidth]{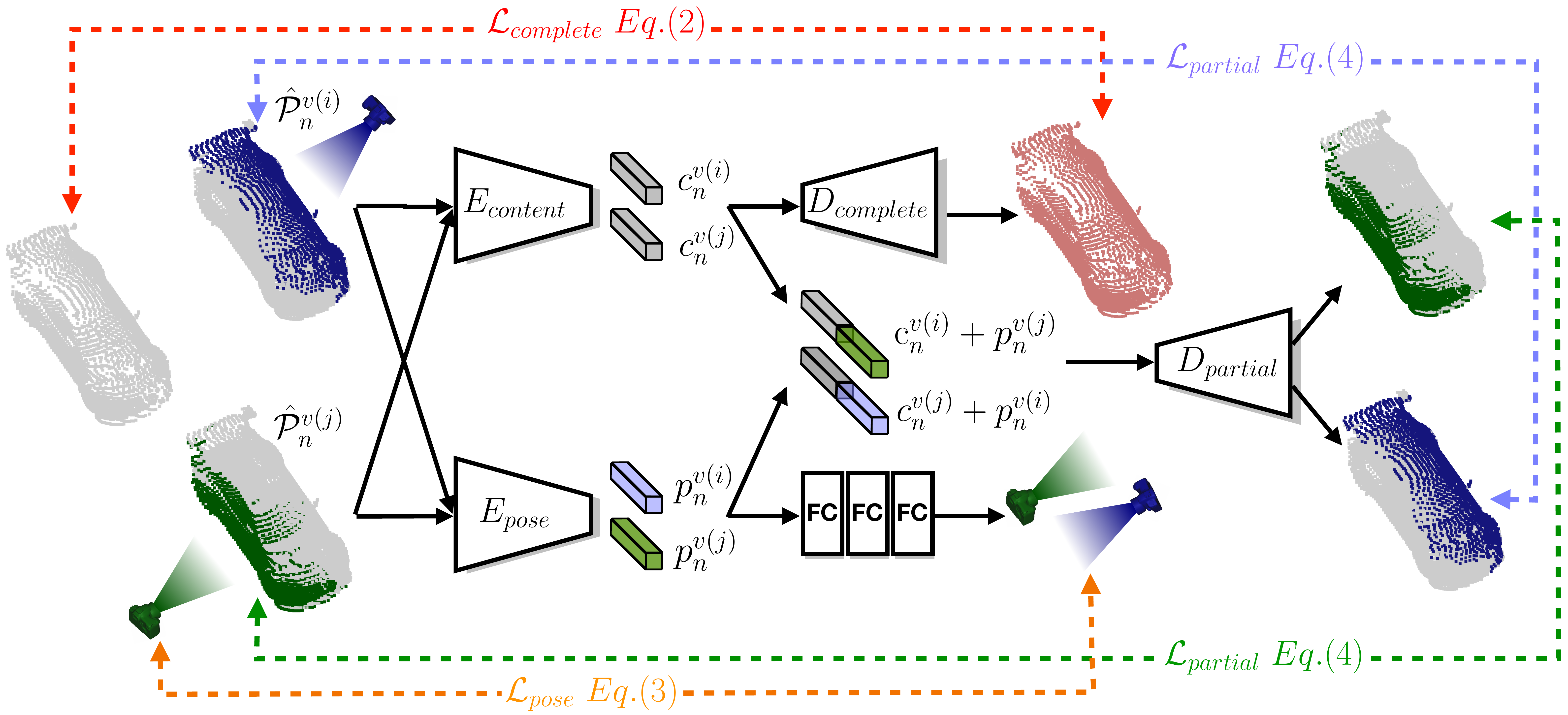}
    \end{center}
    \vspace{-2em}
    \caption{Our proposed framework aims to learn two distinct features ($c_n, p_n$) via disentanglement that consists of three main branches: 1) a completion branch containing a content encoder $E_{content}$ and a decoder $D_{complete}$ that learns the
    content features $c_n$, through completing partial point clouds of the same 3D object. 2) a pose regression branch with a pose encoder $E_{pose}$ that learns the
    pose features $p_n$, through predicting view-angles where a partial point cloud is scanned from. 3) a partial reconstruction branch $D_{partial}$, which leverages combined content features with exchanged pose features to reconstruct the partial inputs. Note that ``$+$'' denotes the concatenation operation and dashed lines in colors are the loss functions we employ for three branches.}
    \label{fig:architecture}
    \vspace{-1em}
\end{figure*}

\vspace{-2mm}
\paragraph{Unsupervised / Self-Supervised Feature Learning \students{for Point Cloud Data.}}
Recently, various works have been proposed to explore the unsupervised and self-supervised schemes for learning feature representations of 3D point clouds, 
where the training objectives or the labels are produced from the data itself.
An intuitive objective is derived via performing the self-reconstruction~\cite{achlioptas2018learning, deng2018ppf}, in which the semantic and compact feature representations are learnt
in the bottleneck between the encoding and decoding processes. For instance, the encoder of 3DCapsuleNet~\cite{zhao20193d} is designed to process 3D point clouds and aggregate the final latent capsules based on the dynamic routing, followed by the decoder to reconstruct the original point cloud composed of multiple point patches.
\walon{\cite{yang2021progressive} proposes an autoencoder architecture where the rich point-wise features in multiple stages are progressively produced through the seed generation module.
Eckart \etal~\cite{eckart2021self} further extend the traditional autoencoder paradigm to form the bottleneck layer being modeled by a discrete generative model.} 
In addition to the self-reconstruction, MAP-VAE~\cite{han2019multi} proposes the multi-angle analysis to introduce the local self-reconstruction, leading a better modeling on the local geometry of point clouds. 

Another effective technique is stemmed from contrastive learning. For instance, PointContrast~\cite{xie2020pointcontrast} proposes to use a pretext task, where the mapping between two point clouds taken by viewing from different perspectives should be consistent even after performing the random geometric transformations on them. \walon{\cite{huang2021spatio} relies on the similar consistency but extends to include the spatial augmentations on the data, and adopts the self-ensembling mechanism to drive the learning.}
Other methods are inspired by the techniques originally applied on 2D images. For instance, Sauder \etal~\cite{sauder2019self} adopts the jigsaw puzzle idea on 3D point cloud, in which the point cloud is decomposed into multiple parts followed by random permutation, and then the network is trained by putting these parts back to their original positions. Similarly, Alliegro \etal~\cite{alliegro2020joint} further integrate multi-task learning into the same framework as \cite{sauder2019self}. 
\walon{In addition, \cite{chen2021shape} destroys shape parts and learns the feature representation via the process of distinguishing the destroyed parts and restoring them.}

Different from the above methods, our approach is inspired by the idea of 2D image inpainting~\cite{pathak2016context} but in a 3D version, in which we propose to self-learn the completion process from partial point clouds as the pretext task. Furthermore, beyond point cloud completion, we extend our framework to consider the pose information, so that features can be disentangled during the completion process to learn effective feature representations. \walon{We note that, a recent work~\cite{wang2021unsupervised} also adopts the idea of point cloud completion to drive feature learning but does not fully leverage the pose information, in which this work can be treated as an ablated variant of our model. Later in experiments we provide the ablation study to verify the benefit of our model design with respect to this work.}

\vspace{-3mm}
\paragraph{Feature Disentanglement.}
Disentanglement aims to decompose the feature representation into multiple parts for better explaining the factors behind data variation. Numerous methods are developed in 2D images, such as the conditional GAN~\cite{mirza2014conditional}, auxiliary classifier GAN~\cite{odena2017conditional}, InfoGAN~\cite{chen2016infogan} and conditional VAE~\cite{sohn2015learning}, all of which are used to disentangle the latent space of GAN~\cite{goodfellow2014generative} and VAE~\cite{kingma2013auto}.
Moreover, introducing disentanglement into representation learning benefits various computer vision applications. For instance, \cite{zou2020joint,li2019cross} disentangle the pose information from person images to obtain the purified feature of person, which improves the performance in the task of person re-identification.
Zhou \etal~\cite{zhou2020unsupervised} factors out the shape and pose features from the 3D mesh data to boost the performance in the tasks of pose transfer and shape retrieval.
In our proposed framework, we disentangle the content and pose information of the 3D point cloud representation, in which the content feature is beneficial for the downstream tasks of classification and part segmentation, while the pose feature boosts the task of point cloud registration.

\section{Proposed Method}
\label{sec:methods}

As motivated in the introduction, the objective of our proposed framework is to perform self-supervised learning via partial point cloud completion and learn the 3D point cloud feature representations, which is decomposed into the content and pose parts. The overall architecture of our proposed framework is illustrated in Figure~\ref{fig:architecture}, where there are three main branches: completion branch, pose regression branch, and partial reconstruction branch. We now detail our proposed framework in the following subsections.

\subsection{Completion Branch}
\label{sec:completion}
The purpose of this branch is to perform 3D partial point cloud completion, which is proposed in this paper as an analogous strategy to 2D image inpainting~\cite{pathak2016context} for self-learning feature representations.
To this end, given a dataset composed of $N$ complete point clouds of 3D objects $\mathbb{P} = \{\mathcal{P}_n\}_{n=1}^N$, for each of the complete point cloud $\mathcal{P}_n$, we are able to generate many of its corresponding partial point clouds $\{\hat{\mathcal{P}}_n^{v(k)}\}_{k=1}^K$ as being scanned from $K$ different pre-defined viewpoints $\{v(k)\}_{k=1}^K$. Then, for an input partial point cloud $\hat{\mathcal{P}}_n^{v(k)}$, we use the content encoder $E_{content}$ to extract its content feature $c_n^{v(k)} = E_{content} (\hat{\mathcal{P}}_n^{v(k)})$, followed by the completion decoder $D_{complete}$ to reconstruct the corresponding complete point cloud $\mathcal{P}_n$ from $c_n^{v(k)}$.

Here, we adopt the standard point cloud feature extractor (e.g., PointNet~\cite{qi2017pointnet} or DGCNN~\cite{dgcnn}) as the content encoder $E_{content}$. For the completion decoder $D_{complete}$, we utilize a morphing-based decoder in MSN~\cite{liu2020morphing} to predict $\mathcal{P}_n$.
To measure the completion quality, we use Earth Mover’s Distance (EMD)~\cite{fan2017point} loss (which is typically approximated via the auction algorithm~\cite{bertsekas1992auction} in practical implementation).
Therefore, given two point clouds $\mathcal{P}_i, \mathcal{P}_j \in \mathbb{R}^3$ with an equal size $\left | \mathcal{P}_i \right | = \left | \mathcal{P}_j \right |$, the EMD loss function is:
\begin{equation}
\label{eq:EMD}
d_{EMD}(\mathcal{P}_i, \mathcal{P}_j) = \min_{\phi (\mathcal{P}_i) \rightarrow \mathcal{P}_j} \sum_{x \in \mathcal{P}_i} \left \| x-\phi (x)) \right \| _2,
\end{equation}
where $\phi$ is an optimal bijection function that allows each point in $\mathcal{P}_i$ to find a nearest corresponding point in $\mathcal{P}_j$.

Moreover, in addition to the EMD loss that aims at guiding the entire generated point cloud to maximally cover the ground truth one, we also adopt the expansion loss proposed in~\cite{liu2020morphing} to better handle local regions in the point cloud. \students{We provide details of this expansion loss in the supplementary material.} As a result, our loss function in the completion branch for a point cloud $\mathcal{P}_n$ in a view-angle $v(k)$ can be written as:
\begin{equation}
\begin{aligned}
\mathcal{L}_{complete} &= d_{EMD}(\mathcal{P}_n , D_{complete}(c_n^{v(k)})) \\ &+ \lambda_{ex}* \mathcal{L}_{expan}(D_{complete}(c_n^{v(k)})).
\label{eq:completion}
\end{aligned}
\end{equation}
\students{where we follow~\cite{liu2020morphing} to set $\lambda_{ex} =0.1$.}

\subsection{Pose Regression Branch}
\label{sec:pose}
Although the self-supervised objective in Section~\ref{sec:completion} enables our completion model to learn feature representations as an initial step, it is still insufficient to fully exploit the rich information contained in partial point cloud captured under various view-angles.
Thus, in the pose regression branch, we aim to learn the pose feature that can predict the pose of the partial data and assist in the feature learning process.
We first use \students{the standard point cloud feature extractor (e.g., PointNet~\cite{qi2017pointnet} or DGCNN~\cite{dgcnn}) as} the pose encoder $E_{pose}$ to extract the pose feature $p_n^{v(k)} = E_{pose} (\hat{\mathcal{P}}_n^{v(k)})$, which represents the feature for the view-angle $v(k)$ of the point cloud $\mathcal{P}_n$. Then, three fully-connected layers are constructed as the pose regressor for predicting the view-angle.

Specifically, our pose regression branch is learnt to predict the camera position where the partial data is scanned from.
We define our camera position in the spherical coordinate system $(\gamma, \theta, \phi)$, where we only rotate the data along the polar angle $\theta$ and the azimuthal angle $\phi$. Note that we fix the radial distance $\gamma$ to have a consistent point cloud density and use the degree as unit of $\theta$ and $\phi$.
Although we pre-define the poses and can consider pose prediction as a classification problem, we find that retaining it as the regression task serves better for feature learning \students{(please refer to our supplementary material)}.
To optimize the pose regression branch, we use the mean square error (MSE). 
Given a partial point cloud $\hat{\mathcal{P}}_n^{v(k)}$, 
our pose regression branch should predict the camera position $\hat{v}(k)$. The loss function thus can be written as:
\begin{equation}
\label{eq:pose}
\mathcal{L}_{pose} = \lVert v(k)-\hat{v}(k) \rVert_2.
\end{equation}

\subsection{Partial Reconstruction Branch}
Based on Section~\ref{sec:completion} and~\ref{sec:pose}, we have constructed two self-supervised objectives, individually for the point cloud completion and the pose regression tasks.
However, there is still a lack of how to make connections between content and pose features, such that the learnt feature representation in each branch is more compact and meaningful.
To achieve it, we propose to disentangle these two factors via the partial reconstruction branch.

Given two partial point clouds $\hat{\mathcal{P}}_n^{v(i)}$ and $\hat{\mathcal{P}}_n^{v(j)}$ derived from the same complete point cloud $\mathcal{P}_n$ but with different viewpoints (i.e., $i \neq j$), 
we hypothesize that the content features from two views should be similar to each other (i.e., $c_n^{v(i)} \sim c_n^{v(j)}$),
as they are from the same 3D object and are used to obtain the same completion output.
As such, if we combine one of the content features with a specific pose feature, this combined feature should follow the pose information to reconstruct the partial point cloud, regardless of which content feature is selected.

To realize this objective, we use a partial decoder $D_{partial}$, which takes the concatenated content and pose features as the input, but with different viewpoints, e.g., concatenation of $c_n^{v(i)}$ and $p_n^{v(j)}$, to reconstruct a partial point cloud $\hat{\mathcal{P}}_n^{v(j)}$. The loss of this partial point cloud reconstruction is based on the EMD loss~\eqref{eq:EMD}:
\begin{equation}
\begin{split}
\label{eq:partial}
    \mathcal{L}_{partial} = d_{EMD}(\hat{\mathcal{P}}_n^{v(i)}, D_{partial}(c_n^{v(j)}, p_n^{v(i)})) \\
                + d_{EMD}(\hat{\mathcal{P}}_n^{v(j)}, D_{partial}(c_n^{v(i)}, p_n^{v(j)})),
\end{split}
\end{equation}
where we consider two terms by exchanging the content and pose features in two viewpoints.

\subsection{Model Pre-training as Pretext}
\label{sec:pretext_data_hyper}

\paragraph{Overall Objective.} Without the use of any annotations, we combine all the aforementioned loss functions from the three branches as our final self-supervised objectives:
\begin{equation}
\label{equ:model}
\mathcal{L}_{all} = \lambda_c \mathcal{L}_{complete} + \lambda_{pa} \mathcal{L}_{partial} + \lambda_{po} \mathcal{L}_{pose},
\end{equation}
where the hyperparameters $\lambda$ control the balance between loss functions, for all the experiments shown later, we set $\lambda_c = \lambda_{pa} = 0.5$ to equally balance two reconstruction-based objectives and set $\lambda_{po} = 0.01$.

\vspace{-3mm}
\paragraph{Dataset, Data Generation, and Implementation Details.}~
We follow the standard self-supervised setting in \cite{sauder2019self,gadelha2020label} and use the ShapeNetCore.v1 dataset~\cite{chang2015shapenet} for the pretext task in model pre-training,
where our learnt content and pose encoders (i.e., $E_{content}$ and $E_{pose}$) will be used to extract the content and pose features from the point cloud data respectively to perform various downstream tasks. In the pretext task, we follow the similar experimental settings as in MSN~\cite{liu2020morphing} for point cloud completion, where we choose a total of $35,827$ 3D models of 8 classes (i.e., table, chair, car, airplane, sofa, lamp, watercraft, and cabinet) from the ShapeNet.
We use Blender software~\cite{Blender} to render the partial point clouds for each of the 3D models as being scanned from 26 pre-defined viewpoints (details are in the supplementary material).
The 3D point coordinates in the point cloud are normalized into a unit sphere (i.e., within $[-1, 1]$). For each point cloud, we uniformly sample 1024 points. The pre-training of our proposed model on the ShapeNet runs for 50 epochs with the Adam optimizer. The learning rate is initialized to 0.001 and decreased by 90\% every 20 epochs, and the batch size is set to 32.

\vspace{-3mm}
\paragraph{Model Architecture.}~In our experiments, we use the standard point cloud feature extractors (e.g., PointNet~\cite{qi2017pointnet} or DGCNN~\cite{dgcnn}) as our encoders, and adopt the morphing-based decoder proposed by MSN~\cite{liu2020morphing} for our decoders. In particular, as the decoder generates the point cloud via composing multiple local patches, we utilize 16 local patches for the completion decoder $D_{complete}$, while the partial decoder $D_{partial}$ only uses a single local patch for partial point cloud reconstruction. 
For both the content encoder $E_{content}$ and pose encoder $E_{pose}$, their architectures are identical to each other but without sharing weights. 
For the pose regressor, we use three fully-connected layers with BatchNorm and ReLU operations between every adjacent layer. More details of the architecture are provided in the supplementary materials. 
Source code and models will be released to the public.

\section{Experimental Results and Analysis}
\label{sec:experiments}
Following the setting in self-supervised point cloud methods~\cite{sauder2019self,gadelha2020label}, we conduct extensive experiments to show the effectiveness and robustness of the point cloud features learnt by our proposed pretext task pre-trained on ShapeNet~\cite{chang2015shapenet}.
Specifically, during training the classifiers of downstream tasks, the pretrained content and pose encoders via our pretext task remain fixed, in which we use the content features for point cloud classification and part segmentation, and the pose features for the registration.
In the following, we first present the analysis of our pretext task via point cloud completion then provide results of each downstream task. 

\subsection{Pretext Analysis}
\begin{figure}[ht]
    \begin{center}
        \includegraphics[width=1\columnwidth]{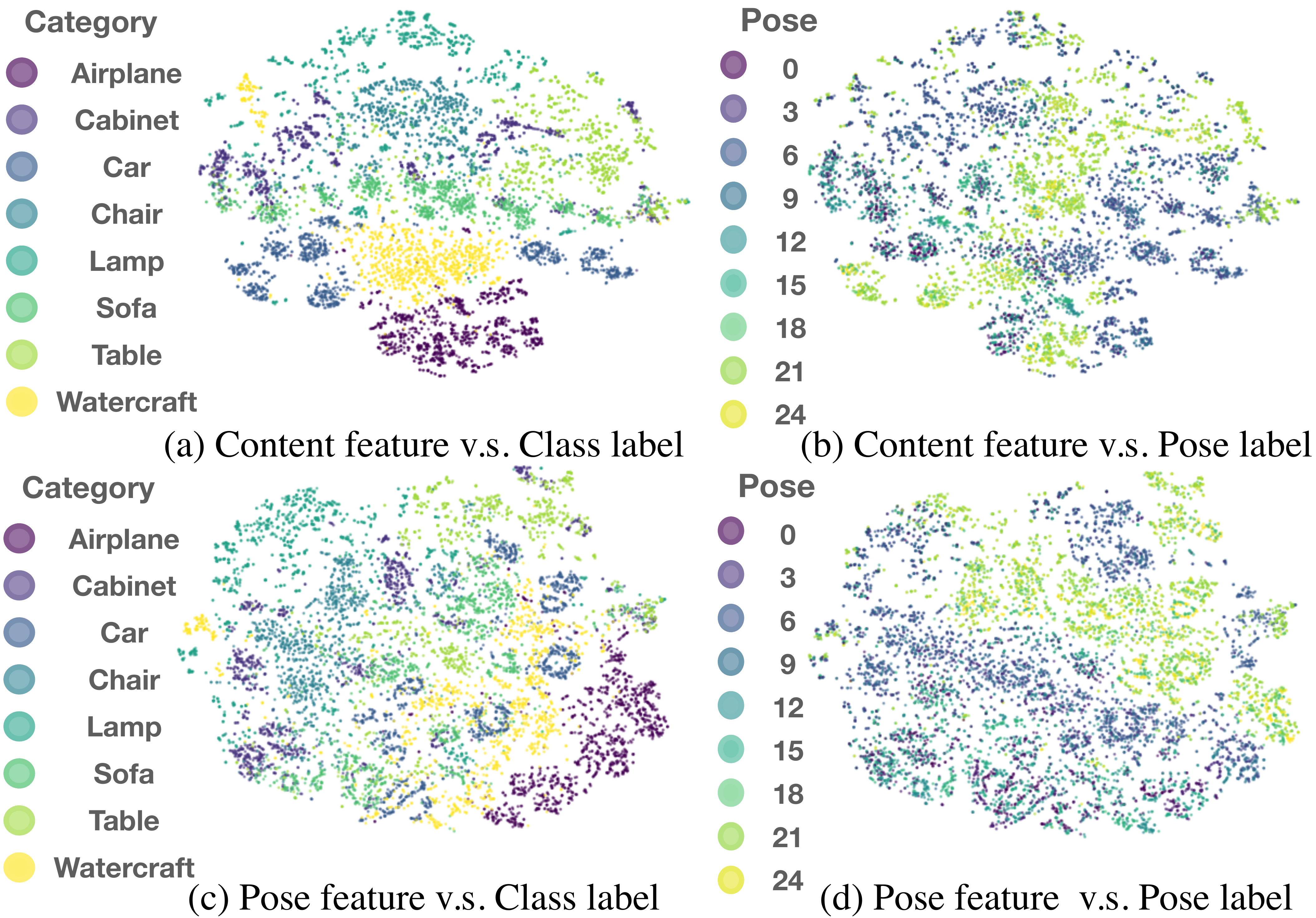}
    \end{center}
    \vspace{-1.5em}
    \caption{\students{t-SNE visualization of content and pose features labeled with class and pose labels respectively on the ShapeNet dataset. The class label indicates the 8 categories data that we used for pre-training, while the pose label indicates the 26 viewpoints that we use for rendering partial data. Note that the pose labels with closer value denote they have similar viewpoints.}
    }
    \label{fig:tSNE}
    \vspace{-1em}
\end{figure}

\paragraph{Features Distribution.}
We assess the quality of our learnt content and pose features by visualizing the feature embedding using t-SNE in Figure~\ref{fig:tSNE}. 
We extract both content and pose features from the fixed $E_{content}$ and $E_{pose}$ respectively, which is pre-trained on ShapeNet using the DGCNN backbone at 50 epoch.
In Figure~\ref{fig:tSNE}{\color{blue}.a} and \ref{fig:tSNE}{\color{blue}.c}, we plot feature distributions with respect to the class label in different colors (note that class labels are not used in training). It shows that data with the same class is grouped together for the content feature, while the pose feature is more invariant to the class label.
Similar observations are also found with respect to the pose label in Figure~\ref{fig:tSNE}{\color{blue}.b} and \ref{fig:tSNE}{\color{blue}.d}, where the content feature is more invariant to the pose label.
This evidence verifies our proposed feature disentanglement process for learning the distinct characteristics of the content and pose features.

\begin{figure}[ht]
    \begin{center}
        \includegraphics[width=1\columnwidth]{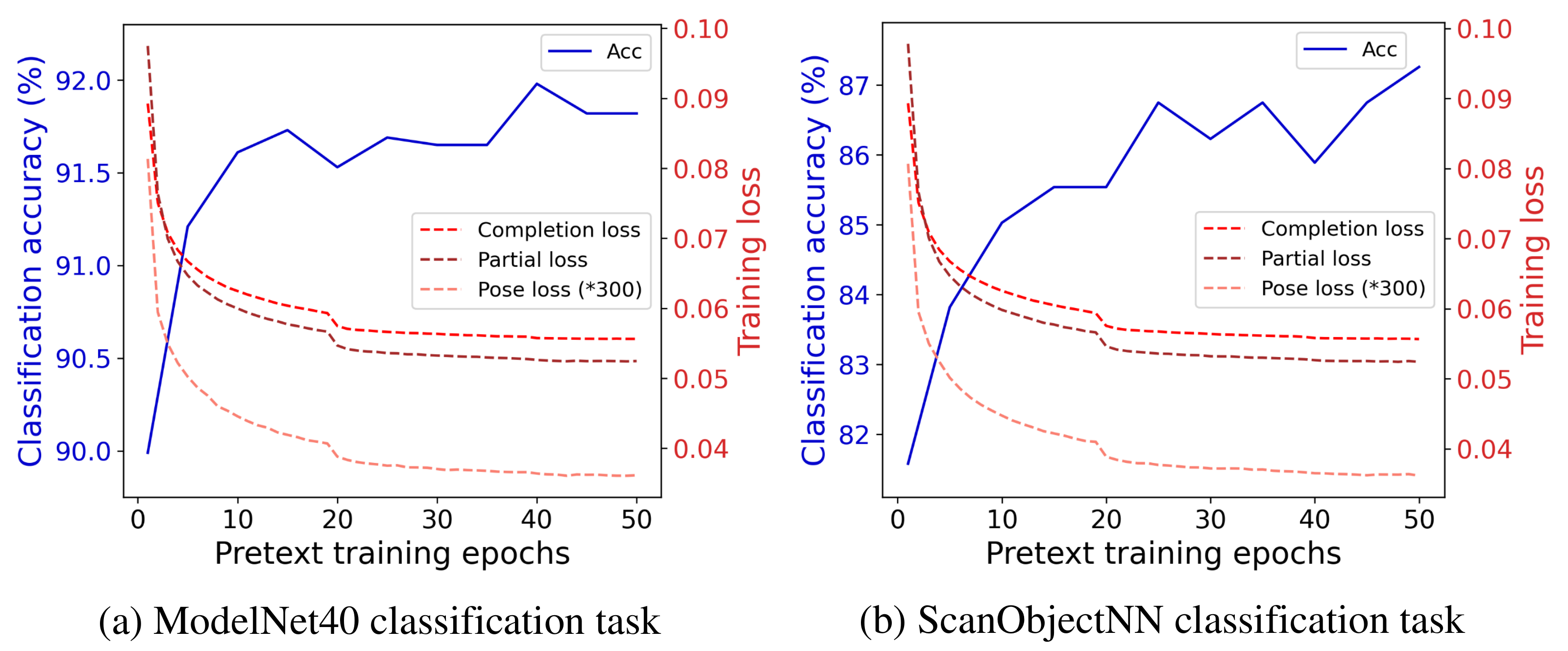}
    \end{center}
    \vspace{-1.5em}
    \caption{\students{Pretext training loss versus downstream classification accuracy. Note that our pretext model is trained on the ShapeNet dataset, while two classification tasks are conducted on the ModelNet40 (a) and ScanObjectNN (b) benchmarks respectively.}
    }
    \label{fig:loss_vs_acc}
    \vspace{-4mm}
\end{figure}

\vspace{-3mm}
\paragraph{Relevance b/w Pretext Loss and Downstream Accuracy.}~\students{We plot the training loss with respect to downstream classification accuracy in Figure~\ref{fig:loss_vs_acc} to verify the effectiveness of our objective function. Both Figure~\ref{fig:loss_vs_acc}{\color{blue}.a} and Figure~\ref{fig:loss_vs_acc}{\color{blue}.b} show the gradual accuracy gain on the downstream tasks as the training loss decreases, which indicates that the optimization of our model is effective in feature learning.
}

\vspace{-1em}

\vspace{3mm}
\subsection{Downstream Task 1: Classification}
\label{exp:cls}
\paragraph{Dataset.}~ModelNet40~\cite{wu20153d} and ScanObjectNN~\cite{angelina2019revisiting} benchmarks are adopted for the downstream task of classification, and make comparisons against several state-of-the-art supervised and unsupervised methods. ModelNet40 is composed of 12,311 CAD models from 40 categories, in which it is split into 9,843 and 2,468 models for training and testing respectively. For the ScanObjectNN dataset, which is a real-world point cloud dataset based on scanned indoor scene data, we use its two variants, ``OBJ\_ONLY'' and ``PB\_T50\_RS'', that are widely adopted for evaluating pre-trained features in the classification downstream tasks~\cite{alliegro2020joint, rao2020global}. ``OBJ\_ONLY'' contains 2,890 models with 15 categories, and ``PB\_T50\_RS'' is stemmed from ``OBJ\_ONLY'' but being augmented with various perturbations (e.g., translation, rotation, scaling, and cropping with background, resulting in total 14,298 models), and hence is considered to be the hardest case in the ScanObjectNN. We follow the default setting to the train/test split for ``OBJ\_ONLY'' and ``PB\_T50\_RS'' by 2,309/581 and 11,416/2,882 respectively.

\vspace{-3mm}
\paragraph{Experimental Setting.}~We follow the standard experimental procedure as in \cite{han2019multi, sauder2019self}, where a linear Support Vector Machine (SVM)~\cite{cortes1995support} is trained on the 3D point cloud features extracted from each of the methods, 
to evaluate the effectiveness of the feature representations in classification (note that our model uses the content feature here).

Our proposed model is pre-trained on ShapeNet, which uses the same pre-trained dataset as the unsupervised learning methods~\cite{yang2018foldingnet, zhao20193d, jing2020self, hassani2019unsupervised, gadelha2020label, sauder2019self} presented in our experiments.
Please note that for both our proposed model and the unsupervised learning methods, the training set of ModelNet40 (or ScanObjectNN) is only used to train the linear SVM classifier.
\students{For fair comparisons with GraphTER~\cite{gao2020graphter} and PointGLR~\cite{rao2020global}, we reproduce their performance by training their pretext model on ShapeNet, where the same reproduction of~\cite{gao2020graphter,rao2020global} is also applied later on other downstream tasks of part segmentation and point cloud registration.
}

\begin{table}[!t]
\centering
\footnotesize
\renewcommand{\arraystretch}{1.1}
\setlength{\tabcolsep}{5pt}
\caption{\textbf{Classification accuracy (\%) on the ModelNet40 dataset.} We compare our proposed model with various state-of-the-art self-supervised methods (denoted as ``SSL''). In the bottom two groups, we compares methods either using the PointNet-based or DGCNN-based backbone. The first two supervised methods are served as references as they fully train the entire network. ($\dagger$: reproduced by training the pretext model on the ShapeNet dataset.)
}
\vspace{-3mm}
\begin{tabular}{lcc}
\toprule
Methods                                     & SSL       & Accuracy      \\
\midrule
PointNet~\cite{qi2017pointnet}              & \xmark    & 89.2          \\
DGCNN~\cite{dgcnn}                          & \xmark    & 92.9          \\
\midrule
GraphTER$^\dagger$~\cite{gao2020graphter}  & \cmark    & 87.8          \\
FoldingNet~\cite{yang2018foldingnet}        & \cmark    & 88.4          \\
PointCapsNet~\cite{zhao20193d}              & \cmark    & 88.9          \\
Multi-Tasks~\cite{hassani2019unsupervised}  & \cmark    & 89.1          \\
\walon{Yang~\etal~\cite{yang2021progressive}} & \cmark    & 90.9          \\
PointGLR$^\dagger$~\cite{rao2020global}     & \cmark    & 91.7          \\
\midrule
Sauder~\etal~\cite{sauder2019self}          & \cmark    & 87.3          \\
ACD~\cite{gadelha2020label} (PointNet++)    & \cmark    & 89.8          \\
\walon{Chen~\etal~\cite{chen2021shape}} & \cmark    & 89.9   \\
PN\_ours                                    & \cmark    & \textbf{90.1}  \\
\midrule
Jing~\etal~\cite{jing2020self}              & \cmark    & 89.8          \\
Sauder~\etal~\cite{sauder2019self}          & \cmark    & 90.6          \\
\walon{STRL~\cite{huang2021spatio}} & \cmark    & 90.9          \\
\walon{ParAE~\cite{eckart2021self}} & \cmark    & 91.6          \\
DGCNN\_ours                                 & \cmark    & \textbf{92.0} \\
\bottomrule
\end{tabular}
\label{tab:ModelNet_cls}
\vspace{-1em}
\end{table}

\begin{table}[!t]
\centering
\footnotesize
\renewcommand{\arraystretch}{1.1}
\setlength{\tabcolsep}{5pt}
\caption{\textbf{Classification accuracy (\%) on the ScanObjectNN dataset.} We compare our proposed model with state-of-the-art self-supervised methods (denoted as ``SSL''), where supervised methods are served as references as they fully train the entire network. ($\dagger$: reproduced by training the pretext model on ShapeNet.)}
\vspace{-3mm}
\begin{tabular}{lccc}
\toprule
Methods                                     & SSL       & OBJ\_ONLY     & PB\_T50\_RS   \\
\midrule
Pointnet~\cite{qi2017pointnet}              & \xmark    & 79.2          & 68.2          \\
DGCNN~\cite{dgcnn}                          & \xmark    & 86.2          & 78.1          \\
\midrule
GraphTER$^\dagger$~\cite{gao2020graphter}  & \cmark    & 72.8          & 60.3          \\
PointGLR$^\dagger$~\cite{rao2020global}     & \cmark    & 85.2          & 73.4          \\
PN\_ours                                    & \cmark    & 84.0          & 70.6          \\
DGCNN\_ours                                 & \cmark    & \textbf{87.3} & \textbf{74.8} \\  
\bottomrule
\end{tabular}
\label{tab:SON_cls}
\vspace{-5mm}
\end{table}

\begin{table*}[!t]
\centering
\footnotesize
\renewcommand{\arraystretch}{1.1}
\setlength{\tabcolsep}{3pt}
\caption{\textbf{Part segmentation results on the ShapeNet-Part dataset.} 
We compare our proposed model with state-of-the-art self-supervised methods (denoted as ``SSL''), where supervised methods are served as references as they fully train the entire network. ($\uppsi$: pretext model is trained on ModelNet40; $*$: reported by MAP~\cite{han2019multi}; \students{$\dagger$: reproduced by training the pretext model on the ShapeNet dataset}.)}
\vspace{-3mm}
\begin{tabular}{lcc|cccccccccccccccc}
\toprule
Methods                                 & SSL    & mIoU & Aero  & Bag & Cap   & Car   & Chair     & Ear ph.   & Guitar    & Knife     & Lamp  & Laptop    & Motor     & Mug  & Pistol & Rocket & Skate & Table \\
\midrule
PointNet~\cite{qi2017pointnet}          & \xmark & 83.7 & 83.4 & 78.7 & 82.5 & 74.9 & 89.6  & 73.0    & 91.5   & 85.9  & 80.8 & 95.3   & 65.2  & 93.0 & 81.2  & 57.9   & 72.8  & 80.6  \\
DGCNN~\cite{dgcnn}                      & \xmark & 85.2 & 84.0 & 83.4 & 86.7 & 77.8 & 90.6 & 74.7 & 91.2 & 87.5 & 82.8 & 95.7 & 66.3 & 94.9 & 81.1 & 63.5 & 74.5 & 82.6 \\
\midrule

Latent-GAN$^*$~\cite{achlioptas2018learning}& \cmark & 57.0 & 54.1 & 48.7 & 62.6 & 43.2 & 68.4 & 58.3 & 74.3 & 68.4 & 53.4 & 82.6 & 18.6 & 75.1 & 54.7 & 37.2 & 46.7 & 66.4 \\
MAP$^\uppsi$~\cite{han2019multi}               & \cmark & 68.0 & 62.7 & 67.1 & 73.0 & 58.5 & 77.1 & 67.3 & 84.8 & 77.1 & 60.9 & 90.8 & 35.8 & 87.7 & 64.2 & 45.0 & 60.4 & 74.8 \\
GraphTER$^\dagger$~\cite{gao2020graphter}       & \cmark & 82.3 & 81.7 & 76.0 & 83.2 & 74.9 & 84.8 & 64.6 & 90.9 & 87.1          & 82.5 & \textbf{95.6} & 56.3 & \textbf{93.2} & 81.6 & \textbf{56.2}          & 71.0 & 67.8 \\ 
PointGLR$^\dagger$~\cite{rao2020global} & \cmark & 84.6 & 81.5 & \textbf{84.7} & 81.7 & 75.4 & \textbf{89.8} & 76.1 & 89.8 & 84.9 & 83.3 & 95.2 & 65.4 & 92.8 & 79.9 & 55.7 & 73.7 & \textbf{83.5} \\

PN\_ours        & \cmark & 83.8 & 81.6 & 73.2 & 83.5 & 74.2 & 89.0 & 70.9 & 89.9 & 85.5 & 80.6 & 95.0 & 66.4 & 92.6 & 82.0          & 53.7 & 72.7 & 82.9 \\
DGCNN\_ours     & \cmark & \textbf{85.1} & \textbf{82.3} & 83.5 & \textbf{84.5} & \textbf{77.3} & \textbf{89.8} & \textbf{76.3} & \textbf{91.0} & \textbf{87.3}          & \textbf{84.2}          & 95.5 & \textbf{67.8}          & 92.5 & \textbf{82.8} & 52.1 & \textbf{73.9} & \textbf{83.5} \\
\bottomrule
\end{tabular}
\label{tab:ShapeNet_part}
\vspace{-3mm}
\end{table*}

\vspace{-3mm}
\paragraph{Results.}~Table~\ref{tab:ModelNet_cls} and Table~\ref{tab:SON_cls} show the results on the ModelNet40 and ScanObjectNN datasets respectively. Please note that, for all tables in this paper, we denote our full model of using PointNet~\cite{qi2017pointnet} and DGCNN~\cite{dgcnn} as the backbone of encoders as ``PN\_ours'' and ``DGCNN\_ours''. 

In Table~\ref{tab:ModelNet_cls}, we show that our proposed model performs favorably against other unsupervised 
methods, thus verifying the robustness of our self-supervised model in learning more holistic and effective feature representations.
Furthermore, the competitive performance (or with a small gap) of our proposed model compared to the supervised baselines shows the practical potential of our 3D point cloud feature representations. 
In Table~\ref{tab:SON_cls}, we show that our proposed model even outperforms all the supervised and self-supervised learning baselines in ``OBJ\_ONLY'', while providing comparable performance in ``PB\_T50\_RS'' with respect to the supervised methods. In particular, such results verify the generalizability of the feature representation learnt by our proposed method across synthetic and real-world datasets.

\subsection{Downstream Task 2: Part Segmentation}
\paragraph{Dataset.}~In addition to having the downstream task of classification for showcasing the capacity of our content features in modeling the holistic point cloud, here we experiment on another task, part segmentation, to investigate how the fine-grained information of local 3D points is maintained in our learnt content features. We adopt the ShapeNet-Part dataset~\cite{yi2016scalable} for performing this evaluation, which is commonly used on the part segmentation task. ShapeNet-Part dataset consists of 16,872 models with 16 categories, with being split into 13,998 and 2,874 for training and testing respectively. Depending on the object category, 3D points are annotated by 2 to 6 part labels, where there are in total 50 distinct labels for the whole dataset.

\vspace{-3mm}
\paragraph{Experimental Setting.}~We follow the same architecture for the part classifier designed for part segmentation task in the PointNet~\cite{qi2017pointnet} framework.
Similar to the classification task, our proposed self-supervised model is first pre-trained on ShapeNet, \students{and the model is fixed as a feature extractor while training a part segmentation classifier.
Following~\cite{qi2017pointnet}, we extract point-wise features of the final convolutional layer (before max-pooling) in the content encoder $E_{content}$ as local features, while using the content features as global features for learning the segmentation part classifier.
} The evaluation metric is based on the average of intersection-over-union (i.e., mIoU).

\vspace{-3mm}
\paragraph{Results.}~Table~\ref{tab:ShapeNet_part} summarizes the evaluation results and the comparison of our proposed model against several methods on the ShapeNet-Part dataset.
Overall, our model provides better performance than the self-supervised methods, in which it demonstrates that the content features learnt via \students{disentangling}
from partial point clouds with different viewpoints are able to benefit the extraction of more discriminative point-wise features.
Moreover, our proposed model is competitive in comparison to the supervised methods that train the entire network on ShapeNet-Part, which shows the robustness of our learnt 3D point cloud features that are used to only train the part classifier.

\subsection{Downstream Task 3: Point Cloud Registration}
\paragraph{Dataset.}~Other than studying the content features in classification and part segmentation, here we adopt a pose-relevant downstream task, i.e., registration, to evaluate the pose features learnt by our proposed model. We conduct experiments for point cloud registration based on ModelNet40. We follow the same procedure of data preparation as in DCP~\cite{wang2019deep}, in which the whole ModelNet40 dataset is randomly split into training and testing sets regardless of object categories. For each point cloud $\mathcal{P}$, a randomly-drawn rigid transformation $\mathcal{T}$ is applied on $\mathcal{P}$ to obtain the transformed point cloud $\mathcal{T}(\mathcal{P})$, where $\{\mathcal{P}, \mathcal{T}(\mathcal{P})\}$ together with $\mathcal{T}$ becomes an input-output pair for learning the registration model.

\vspace{-3mm}
\paragraph{Experimental Setting.}~We adopt the same architecture for the registration model and follow the training procedure proposed by DCP~\cite{wang2019deep} which uses DGCNN~\cite{dgcnn} as their feature extractor. Our proposed model is first pre-trained on ShapeNet, and then we replace the feature extractor in DCP with our pre-trained pose encoder $E_{pose}$ to train the registration task. Note that $E_{pose}$ is fixed during the training process to verify the effectiveness of our learnt features.
\students{For evaluation, we focus on estimating the orientation/rotation between point clouds in this experiments and adopt the root mean squared error (RMSE) as the evaluation metric.}

\vspace{-3mm}
\paragraph{Results.}~\students{As shown in Table~\ref{tab:ModelNet_regristration}, the pose features learnt from our model pre-trained on the ShapeNet dataset outperform other self-supervised methods on RMSE(R). 
Moreover, our proposed model is competitive against the state-of-the-art supervised methods that fully train the entire network on ModelNet40, which verifies the robustness of our pose features on the point cloud registration task.} 

\begin{table}[!t]
\centering
\footnotesize
\renewcommand{\arraystretch}{1.1}
\setlength{\tabcolsep}{5pt}
\caption{\textbf{Registration results on the ModelNet40 dataset.} \students{We compare our proposed model with supervised methods reported by DCP~\cite{wang2019deep} and self-supervised methods reproduced by using the official implementations.} (
$\dagger$: reproduced by training the pretext model on the ShapeNet dataset.)
}
\vspace{-3mm}
\begin{tabular}{lccc}
\toprule
Methods                                     & SSL       & RMSE(R) $\downarrow$  \\ %
\midrule
PointNetLK~\cite{aoki2019pointnetlk}        & \xmark    & 15.095                \\ %
FGR~\cite{benjemaa1999fast}                 & \xmark    & 9.363                 \\ %
DCP-v2~\cite{wang2019deep} (PN)             & \xmark    & 7.061                 \\ %
DCP-v2~\cite{wang2019deep} (DGCNN)          & \xmark    & 1.143                 \\ %
\midrule
PointGLR$^\dagger$~\cite{rao2020global}     & \cmark    & 3.900                 \\ %
GraphTER$^\dagger$~\cite{gao2020graphter}   & \cmark    & 2.927                 \\ %
PN\_ours                                    & \cmark    & 6.719                 \\ %
DGCNN\_ours                                 & \cmark    & \textbf{1.557}        \\ %
\bottomrule
\end{tabular}
\label{tab:ModelNet_regristration}
\vspace{-4mm}
\end{table}

\subsection{Ablation Study}\label{sec:ablation_study}
\paragraph{Comparisons between Content and Pose Features.}~In the previous experimental results, we adopt the content feature for the tasks of classification and part segmentation, while the pose feature is utilized for the registration. Here we provide results of using different features for these downstream tasks. Table~\ref{tab:compare_cont_pose_cls_part} provides the results of applying either the content or pose feature on all the downstream tasks.

We observe that, in comparison to the pose feature, the content feature works best for the classification and part segmentation tasks, where the former needs more semantic and holistic understanding of the point clouds while the latter needs fine-grained modeling on the local geometric structures. This verifies that the content feature learnt by our proposed method is able to compactly model both the global and local structural information of the input point cloud. When it comes to the downstream task of registration, the pose feature instead contributes better than the content feature. Such results are also aligned with the findings from several prior works~\cite{tsai2017indoor, phuc2017registration, aoki2019pointnetlk}, where the camera pose estimation task could provide benefits for the registration problem.

\begin{table}[!h]
\centering
\footnotesize
\renewcommand{\arraystretch}{1.1}
\setlength{\tabcolsep}{5pt}
\caption{\textbf{Content v.s. pose features.} Comparisons between content and pose features in \students{the classification, part segmentation and registration tasks. (MN40: ModelNet40; SON: we use ``OBJ\_ONLY'' on ScanObjectNN; Regis.: we use RMSE(R) as evaluation metric.)}
}
\vspace{-3mm}
\begin{tabular}{cl|cc|c|c}
\toprule
\multirow{2}{*}{Backbone}   & \multirow{2}{*}{Feature}  & \multicolumn{2}{c|}{Classification}   & \multirow{2}{*}{Part Seg.}    & \multirow{2}{*}{Regis.}   \\
                            &                           & MN40              & SON               &                                                           \\ \midrule
\multirow{2}{*}{PN}         & content                   & \textbf{90.11}    & \textbf{83.99}    & \textbf{83.83}                & 6.875                     \\
                            & pose                      & 88.98             & 78.49             & 83.54                         & \textbf{6.719}            \\ \midrule
\multirow{2}{*}{DGCNN}      & content                   & \textbf{91.98}    & \textbf{87.26}    & \textbf{85.05}                & 2.920                     \\
                            & pose                      & 90.36             & 82.79             & 84.46                         & \textbf{1.557}            \\ \bottomrule
\end{tabular}
\label{tab:compare_cont_pose_cls_part}
\vspace{-2mm}
\end{table}

\paragraph{Comparisons among Different Model Designs.}~We study different model designs here. In addition to our full model, \students{
we experiment other designs: 1) the framework of only using the completion branch \walon{as proposed by the work of~\cite{wang2021unsupervised}} (denoted as ``Comp-only'');} 2) the framework of only using the pose regression branch (denoted as ``PR-only''); 3) the joint learning framework (denoted as ``JL''), where the completion and pose regression branches share one encoder trained in a multi-tasking manner. More architecture details of these model variants are provided in the supplementary material. The quantitative comparison among these designs is provided in Table~\ref{tab:compare_archi}.

\students{As we expected, Comp-only can learn more semantic information and obtain better results than PR-only in the classification and part segmentation tasks, while PR-only learns more camera related information and obtains better results than Comp-only in the registration task. Furthermore, our full model can extract more robust and effective features than the other designs, especially in the real-world data classification task. Interestingly, JL obtains worse results than PR-only in the classification task. We hypothesize that, as completion and pose regression have quite different characteristics, having one encoder to jointly learn both tasks may result in worse feature learning, thus verifying again the contribution of our proposed disentanglement method.}

\begin{table}[!h]
\centering
\footnotesize
\renewcommand{\arraystretch}{1.1}
\setlength{\tabcolsep}{5pt}
\caption{\textbf{\students{
Comparisons with different model designs.}}
Note that \students{``Comp-only'' only uses the completion branch,} ``PR-only'' only adopts the pose regression branch, and ``JL'' indicates the joint learning manner with the shared encoder.
\students{(MN40: ModelNet40; SON: we use ``OBJ\_ONLY'' on ScanObjectNN.)}}
\vspace{-3mm}
\begin{tabular}{cl|cc|c|c}
\toprule
\multirow{2}{*}{Backbone}   & \multirow{2}{*}{Methods}  & \multicolumn{2}{c|}{Classification}   & \multirow{2}{*}{Part Seg.}    & \multirow{2}{*}{Regis.}   \\
                            &                           & MN40              & SON               &                                                           \\ \midrule
\multirow{4}{*}{PN}         & Comp-only                 & 89.71             & 83.13             & 83.72                         & 7.858                     \\
                            & PR-only                   & 89.42             & 79.00             & 83.60                         & 6.865                     \\
                            & JL                        & 89.34             & 78.49             & 83.77                         & 6.843                     \\
                            & Ours full                 & \textbf{90.11}    & \textbf{83.99}    & \textbf{83.83}                & \textbf{6.719}            \\ \midrule
\multirow{4}{*}{DGCNN}      & Comp-only                 & 91.37             & 86.06             & 84.93                         & 3.231                     \\
                            & PR-only                   & 90.64             & 82.62             & 84.47                         & 1.644                     \\ 
                            & JL                        & 90.52             & 81.93             & 84.49                         & 1.741                     \\
                            & Ours full                 & \textbf{91.98}    & \textbf{87.26}    & \textbf{85.05}                & \textbf{1.557}            \\ \bottomrule
\end{tabular}
\label{tab:compare_archi}
\vspace{-2mm}
\end{table}

\vspace{-2mm}
\section{Conclusions}
\label{sec:conclusion}
In this paper, we propose a self-supervised framework for learning feature representations of 3D point clouds. Based on the objectives composed of completion, reconstruction, and pose regression for the partial point cloud data, our model learns to disentangle the content and pose factors. Our learnt content and pose feature representations of 3D point clouds experimentally demonstrate the superior performance in comparison to other self-supervised methods in various downstream tasks such as classification, part segmentation, and registration.

{\small
\bibliographystyle{ieee_fullname}
\bibliography{egbib}

\begin{thebibliography}{10}\itemsep=-1pt

\bibitem{achlioptas2018learning}
Panos Achlioptas, Olga Diamanti, Ioannis Mitliagkas, and Leonidas Guibas.
\newblock Learning representations and generative models for 3d point clouds.
\newblock In {\em International Conference on Machine Learning (ICML)}, 2018.

\bibitem{alliegro2020joint}
Antonio Alliegro, Davide Boscaini, and Tatiana Tommasi.
\newblock Joint supervised and self-supervised learning for 3d real-world
  challenges.
\newblock {\em ArXiv:2004.07392}, 2020.

\bibitem{angelina2019revisiting}
Mikaela Angelina~Uy, Quang-Hieu Pham, Binh-Son Hua, Duc Thanh~Nguyen, and
  Sai-Kit Yeung.
\newblock Revisiting point cloud classification: A new benchmark dataset and
  classification model on real-world data.
\newblock In {\em IEEE International Conference on Computer Vision (ICCV)},
  2019.

\bibitem{aoki2019pointnetlk}
Yasuhiro Aoki, Hunter Goforth, Rangaprasad~Arun Srivatsan, and Simon Lucey.
\newblock Pointnetlk: Robust \& efficient point cloud registration using
  pointnet.
\newblock In {\em IEEE Conference on Computer Vision and Pattern Recognition
  (CVPR)}, 2019.

\bibitem{benjemaa1999fast}
Raouf Benjemaa and Francis Schmitt.
\newblock Fast global registration of 3d sampled surfaces using a
  multi-z-buffer technique.
\newblock {\em Image and Vision Computing}, 1999.

\bibitem{bertsekas1992auction}
Dimitri~P Bertsekas.
\newblock Auction algorithms for network flow problems: A tutorial
  introduction.
\newblock {\em Computational optimization and applications}, 1992.

\bibitem{chang2015shapenet}
Angel~X Chang, Thomas Funkhouser, Leonidas Guibas, Pat Hanrahan, Qixing Huang,
  Zimo Li, Silvio Savarese, Manolis Savva, Shuran Song, Hao Su, et~al.
\newblock Shapenet: An information-rich 3d model repository.
\newblock {\em ArXiv:1512.03012}, 2015.

\bibitem{chen2016infogan}
Xi Chen, Yan Duan, Rein Houthooft, John Schulman, Ilya Sutskever, and Pieter
  Abbeel.
\newblock Infogan: Interpretable representation learning by information
  maximizing generative adversarial nets.
\newblock In {\em Advances in Neural Information Processing Systems (NeurIPS)},
  2016.

\bibitem{chen2021shape}
Ye Chen, Jinxian Liu, Bingbing Ni, Hang Wang, Jiancheng Yang, Ning Liu, Teng
  Li, and Qi Tian.
\newblock Shape self-correction for unsupervised point cloud understanding.
\newblock In {\em IEEE International Conference on Computer Vision (ICCV)},
  2021.

\bibitem{Blender}
Blender~Online Community.
\newblock {\em Blender - a 3D modelling and rendering package}.
\newblock Blender Foundation, 2018.

\bibitem{cortes1995support}
Corinna Cortes and Vladimir Vapnik.
\newblock Support-vector networks.
\newblock {\em Machine learning}, 1995.

\bibitem{deng2018ppf}
Haowen Deng, Tolga Birdal, and Slobodan Ilic.
\newblock Ppf-foldnet: Unsupervised learning of rotation invariant 3d local
  descriptors.
\newblock In {\em European Conference on Computer Vision (ECCV)}, 2018.

\bibitem{eckart2021self}
Benjamin Eckart, Wentao Yuan, Chao Liu, and Jan Kautz.
\newblock Self-supervised learning on 3d point clouds by learning discrete
  generative models.
\newblock In {\em IEEE Conference on Computer Vision and Pattern Recognition
  (CVPR)}, 2021.

\bibitem{fan2017point}
Haoqiang Fan, Hao Su, and Leonidas~J Guibas.
\newblock A point set generation network for 3d object reconstruction from a
  single image.
\newblock In {\em IEEE Conference on Computer Vision and Pattern Recognition
  (CVPR)}, 2017.

\bibitem{gadelha2020label}
Matheus Gadelha, Aruni RoyChowdhury, Gopal Sharma, Evangelos Kalogerakis,
  Liangliang Cao, Erik Learned-Miller, Rui Wang, and Subhransu Maji.
\newblock Label-efficient learning on point clouds using approximate convex
  decompositions.
\newblock In {\em European Conference on Computer Vision (ECCV)}, 2020.

\bibitem{gao2020graphter}
Xiang Gao, Wei Hu, and Guo-Jun Qi.
\newblock Graphter: Unsupervised learning of graph transformation equivariant
  representations via auto-encoding node-wise transformations.
\newblock In {\em IEEE Conference on Computer Vision and Pattern Recognition
  (CVPR)}, 2020.

\bibitem{goodfellow2014generative}
Ian Goodfellow, Jean Pouget-Abadie, Mehdi Mirza, Bing Xu, David Warde-Farley,
  Sherjil Ozair, Aaron Courville, and Yoshua Bengio.
\newblock Generative adversarial nets.
\newblock In {\em Advances in Neural Information Processing Systems (NeurIPS)},
  2014.

\bibitem{han2019multi}
Zhizhong Han, Xiyang Wang, Yu-Shen Liu, and Matthias Zwicker.
\newblock Multi-angle point cloud-vae: unsupervised feature learning for 3d
  point clouds from multiple angles by joint self-reconstruction and
  half-to-half prediction.
\newblock In {\em IEEE International Conference on Computer Vision (ICCV)},
  2019.

\bibitem{hassani2019unsupervised}
Kaveh Hassani and Mike Haley.
\newblock Unsupervised multi-task feature learning on point clouds.
\newblock In {\em IEEE International Conference on Computer Vision (ICCV)},
  2019.

\bibitem{huang2021spatio}
Siyuan Huang, Yichen Xie, Song-Chun Zhu, and Yixin Zhu.
\newblock Spatio-temporal self-supervised representation learning for 3d point
  clouds.
\newblock In {\em IEEE International Conference on Computer Vision (ICCV)},
  2021.

\bibitem{jing2020self}
Longlong Jing, Yucheng Chen, Ling Zhang, Mingyi He, and Yingli Tian.
\newblock Self-supervised feature learning by cross-modality and cross-view
  correspondences.
\newblock {\em ArXiv:2004.05749}, 2020.

\bibitem{kingma2013auto}
Diederik~P Kingma and Max Welling.
\newblock Auto-encoding variational bayes.
\newblock In {\em International Conference on Learning Representations (ICLR)},
  2014.

\bibitem{kipf2016semi}
Thomas~N Kipf and Max Welling.
\newblock Semi-supervised classification with graph convolutional networks.
\newblock In {\em International Conference on Learning Representations (ICLR)},
  2016.

\bibitem{li2018pointcnn}
Yangyan Li, Rui Bu, Mingchao Sun, Wei Wu, Xinhan Di, and Baoquan Chen.
\newblock Pointcnn: Convolution on x-transformed points.
\newblock In {\em Advances in Neural Information Processing Systems (NeurIPS)},
  2018.

\bibitem{li2019cross}
Yu-Jhe Li, Ci-Siang Lin, Yan-Bo Lin, and Yu-Chiang~Frank Wang.
\newblock Cross-dataset person re-identification via unsupervised pose
  disentanglement and adaptation.
\newblock In {\em IEEE International Conference on Computer Vision (ICCV)},
  2019.

\bibitem{liu2020morphing}
Minghua Liu, Lu Sheng, Sheng Yang, Jing Shao, and Shi-Min Hu.
\newblock Morphing and sampling network for dense point cloud completion.
\newblock In {\em AAAI Conference on Artificial Intelligence (AAAI)}, 2020.

\bibitem{liu2019relation}
Yongcheng Liu, Bin Fan, Shiming Xiang, and Chunhong Pan.
\newblock Relation-shape convolutional neural network for point cloud analysis.
\newblock In {\em IEEE Conference on Computer Vision and Pattern Recognition
  (CVPR)}, 2019.

\bibitem{mirza2014conditional}
Mehdi Mirza and Simon Osindero.
\newblock Conditional generative adversarial nets.
\newblock {\em ArXiv:1411.1784}, 2014.

\bibitem{odena2017conditional}
Augustus Odena, Christopher Olah, and Jonathon Shlens.
\newblock Conditional image synthesis with auxiliary classifier gans.
\newblock In {\em International Conference on Machine Learning (ICML)}, 2017.

\bibitem{pathak2016context}
Deepak Pathak, Philipp Krahenbuhl, Jeff Donahue, Trevor Darrell, and Alexei~A
  Efros.
\newblock Context encoders: Feature learning by inpainting.
\newblock In {\em IEEE Conference on Computer Vision and Pattern Recognition
  (CVPR)}, 2016.

\bibitem{phuc2017registration}
Trong Phuc~Truong, Masahiro Yamaguchi, Shohei Mori, Vincent Nozick, and Hideo
  Saito.
\newblock Registration of rgb and thermal point clouds generated by structure
  from motion.
\newblock In {\em IEEE International Conference on Computer Vision Workshops
  (ICCV Workshops)}, 2017.

\bibitem{qi2017pointnet}
Charles~R Qi, Hao Su, Kaichun Mo, and Leonidas~J Guibas.
\newblock Pointnet: Deep learning on point sets for 3d classification and
  segmentation.
\newblock In {\em IEEE Conference on Computer Vision and Pattern Recognition
  (CVPR)}, 2017.

\bibitem{qi2017pointnet++}
Charles~Ruizhongtai Qi, Li Yi, Hao Su, and Leonidas~J Guibas.
\newblock Pointnet++: Deep hierarchical feature learning on point sets in a
  metric space.
\newblock In {\em Advances in Neural Information Processing Systems (NeurIPS)},
  2017.

\bibitem{rao2020global}
Yongming Rao, Jiwen Lu, and Jie Zhou.
\newblock Global-local bidirectional reasoning for unsupervised representation
  learning of 3d point clouds.
\newblock In {\em IEEE Conference on Computer Vision and Pattern Recognition
  (CVPR)}, 2020.

\bibitem{sauder2019self}
Jonathan Sauder and Bjarne Sievers.
\newblock Self-supervised deep learning on point clouds by reconstructing
  space.
\newblock In {\em Advances in Neural Information Processing Systems (NeurIPS)},
  2019.

\bibitem{sohn2015learning}
Kihyuk Sohn, Honglak Lee, and Xinchen Yan.
\newblock Learning structured output representation using deep conditional
  generative models.
\newblock In {\em Advances in Neural Information Processing Systems (NeurIPS)},
  2015.

\bibitem{tsai2017indoor}
Chi-Yi Tsai and Chih-Hung Huang.
\newblock Indoor scene point cloud registration algorithm based on rgb-d camera
  calibration.
\newblock {\em Sensors}, 2017.

\bibitem{vaswani2017attention}
Ashish Vaswani, Noam Shazeer, Niki Parmar, Jakob Uszkoreit, Llion Jones,
  Aidan~N Gomez, {\L}ukasz Kaiser, and Illia Polosukhin.
\newblock Attention is all you need.
\newblock In {\em Advances in Neural Information Processing Systems (NeurIPS)},
  2017.

\bibitem{wang2021unsupervised}
Hanchen Wang, Qi Liu, Xiangyu Yue, Joan Lasenby, and Matt~J Kusner.
\newblock Unsupervised point cloud pre-training via occlusion completion.
\newblock In {\em IEEE International Conference on Computer Vision (ICCV)},
  2021.

\bibitem{wang2019deep}
Yue Wang and Justin~M Solomon.
\newblock Deep closest point: Learning representations for point cloud
  registration.
\newblock In {\em IEEE International Conference on Computer Vision (ICCV)},
  2019.

\bibitem{dgcnn}
Yue Wang, Yongbin Sun, Ziwei Liu, Sanjay~E. Sarma, Michael~M. Bronstein, and
  Justin~M. Solomon.
\newblock Dynamic graph cnn for learning on point clouds.
\newblock {\em ACM Transactions on Graphics (TOG)}, 2019.

\bibitem{wu20153d}
Zhirong Wu, Shuran Song, Aditya Khosla, Fisher Yu, Linguang Zhang, Xiaoou Tang,
  and Jianxiong Xiao.
\newblock 3d shapenets: A deep representation for volumetric shapes.
\newblock In {\em IEEE Conference on Computer Vision and Pattern Recognition
  (CVPR)}, 2015.

\bibitem{xie2020pointcontrast}
Saining Xie, Jiatao Gu, Demi Guo, Charles~R Qi, Leonidas~J Guibas, and Or
  Litany.
\newblock Pointcontrast: Unsupervised pre-training for 3d point cloud
  understanding.
\newblock In {\em European Conference on Computer Vision (ECCV)}, 2020.

\bibitem{xu2018spidercnn}
Yifan Xu, Tianqi Fan, Mingye Xu, Long Zeng, and Yu Qiao.
\newblock Spidercnn: Deep learning on point sets with parameterized
  convolutional filters.
\newblock In {\em European Conference on Computer Vision (ECCV)}, 2018.

\bibitem{yang2021progressive}
Juyoung Yang, Pyunghwan Ahn, Doyeon Kim, Haeil Lee, and Junmo Kim.
\newblock Progressive seed generation auto-encoder for unsupervised point cloud
  learning.
\newblock In {\em IEEE International Conference on Computer Vision (ICCV)},
  2021.

\bibitem{yang2018foldingnet}
Yaoqing Yang, Chen Feng, Yiru Shen, and Dong Tian.
\newblock Foldingnet: Point cloud auto-encoder via deep grid deformation.
\newblock In {\em IEEE Conference on Computer Vision and Pattern Recognition
  (CVPR)}, 2018.

\bibitem{yi2016scalable}
Li Yi, Vladimir~G Kim, Duygu Ceylan, I-Chao Shen, Mengyan Yan, Hao Su, Cewu Lu,
  Qixing Huang, Alla Sheffer, and Leonidas Guibas.
\newblock A scalable active framework for region annotation in 3d shape
  collections.
\newblock {\em SIGGRAPH Asia}, 2016.

\bibitem{yuan2018pcn}
Wentao Yuan, Tejas Khot, David Held, Christoph Mertz, and Martial Hebert.
\newblock Pcn: Point completion network.
\newblock In {\em International Conference on 3D Vision (3DV)}, 2018.

\bibitem{zhao20193d}
Yongheng Zhao, Tolga Birdal, Haowen Deng, and Federico Tombari.
\newblock 3d point capsule networks.
\newblock In {\em IEEE Conference on Computer Vision and Pattern Recognition
  (CVPR)}, 2019.

\bibitem{zhou2020unsupervised}
Keyang Zhou, Bharat~Lal Bhatnagar, and Gerard Pons-Moll.
\newblock Unsupervised shape and pose disentanglement for 3d meshes.
\newblock In {\em European Conference on Computer Vision (ECCV)}, 2020.

\bibitem{zou2020joint}
Yang Zou, Xiaodong Yang, Zhiding Yu, BVK Kumar, and Jan Kautz.
\newblock Joint disentangling and adaptation for cross-domain person
  re-identification.
\newblock In {\em European Conference on Computer Vision (ECCV)}, 2020.

\end{thebibliography}
}

\end{document}